\documentclass[letterpaper]{article} 
\usepackage{aaai2026}  
\usepackage{times}  
\usepackage{helvet}  
\usepackage{courier}  
\usepackage[hyphens]{url}  
\usepackage{graphicx} 
\usepackage{natbib}  
\usepackage{caption} 
\usepackage{algorithm}
\usepackage{algorithmic}

\usepackage{newfloat}
\usepackage{listings}
\DeclareCaptionStyle{ruled}{labelfont=normalfont,labelsep=colon,strut=off} 
\floatstyle{ruled}
\newfloat{listing}{tb}{lst}{}
\floatname{listing}{Listing}
\usepackage{booktabs}
\usepackage{amssymb}
\usepackage{algorithm}
\usepackage{algorithmic}
\usepackage{amsmath}
\usepackage{amssymb}

\title{An Efficient and Harmonized Framework for Balanced Cross-Domain Feature Integration}
\author{
Shaoxu Li, Ye Pan\thanks{Corresponding author}
}
\affiliations{
John Hopcroft Center for Computer Science, Shanghai Jiao Tong University\\
{lishaoxu, whitneypanye}@sjtu.edu.cn
}

\usepackage{bibentry}

\begin{document}

\twocolumn[{%
\renewcommand\twocolumn[1][]{#1}%
\maketitle
\begin{center}
    \centering
    \captionsetup{type=figure}
    \includegraphics[width=\textwidth]{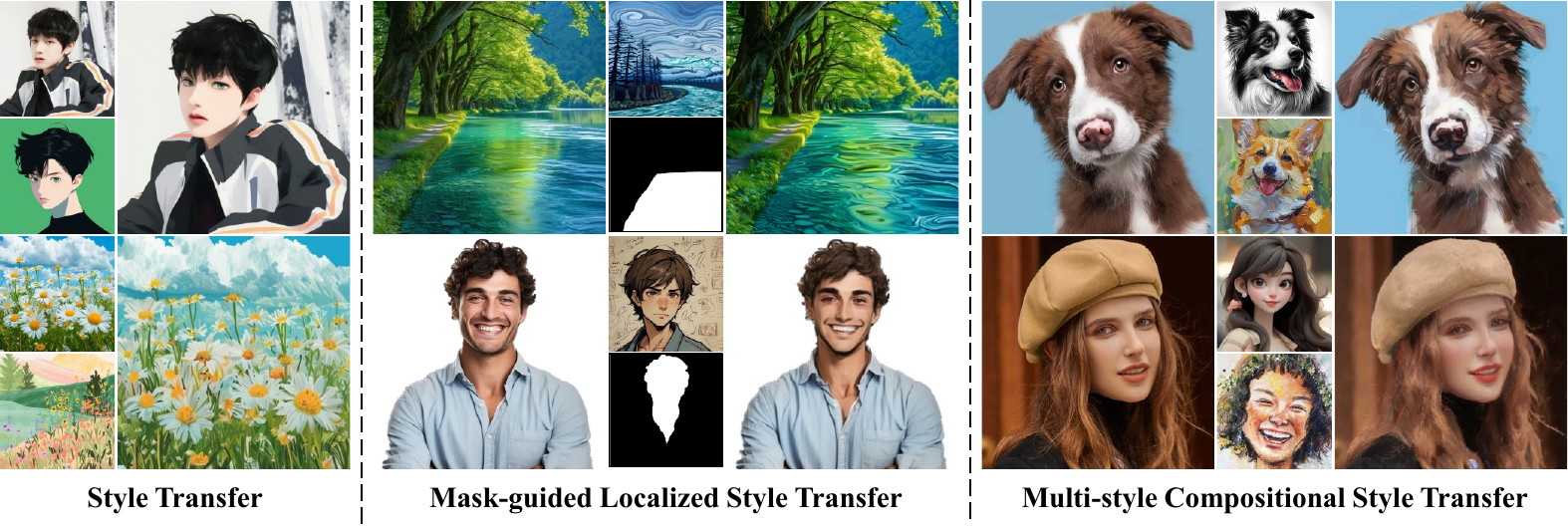}
    \captionof{figure}{Style transfer results using our proposed architecture. By leveraging a content image and a style image as input, our method achieves precise attribute-level style propagation while preserving high-fidelity semantic content integrity. Beyond global style transfer, our approach enables two extensions: mask-guided localized and multi-style compositional style transfer.}
    \label{fig:teaser}
\end{center}%
}]

\begin{abstract}
Despite significant advancements in image generation using advanced generative frameworks, cross-image integration of content and style remains a key challenge. Current generative models, while powerful, frequently depend on vague textual prompts to define styles—creating difficulties in balancing content semantics and style preservation.
We propose a novel framework that utilizes customized models to learn style representations. It enhances content preservation through cross-model feature and attention modulation, leveraging the inherent semantic consistency across models. Additionally, we introduce fixed feature and adaptive attention fusion to achieve the desired balance between content and style. We further develop spatial (mask-guided localized) and temporal (multi-style compositional) multi-model combinations, enabling flexible fusion of models and styles. Extensive experiments demonstrate that our method outperforms state-of-the-art approaches in balancing content preservation and stylistic coherence.
\end{abstract}

\begin{links}
    \link{Code}{https://github.com/lishaoxu1994/DiffStyler}
\end{links}

\section{Introduction}
Paintings, as potent artistic expressions, communicate distinctive perspectives via stylistic components, textures, and motifs. However, their subtle aesthetic qualities remain challenging to articulate in words—a hurdle that poses difficulties for example-based artistic style transfer.
Image style transfer, which integrates artistic styles with content, has garnered interdisciplinary attention\cite{Bin2004Efficient, Gatys_2015_Neural, Zhang2023Inversion, lv2024color, zhou2024comprehensive}. Text-driven approaches, particularly those leveraging diffusion models\cite{song2022denoising, ramesh2022hierarchical}, have propelled the field forward. Yet, they struggle with formulating detailed prompts to capture the essence of a style. Some methods attempt texture inversion\cite{gal2022textual, jeong2024visual} to depict styles. Nevertheless, these methods are often time-consuming and fail to accurately represent the attributes, impeding their practical application.

To address this, we present a diffusion-based framework that learns styles from a single image (e.g., cartoons, paintings) and applies them to natural images. It supports mask-guided localized style transfer and multi-style compositional transfer, offering exceptional control with computational efficiency. As shown in Figure \ref{fig:teaser}, our method enables region-specific transfer via masks and simultaneous multi-style transfer for creative flexibility, generating cohesive outputs that blend content with target styles.  
Building upon text-to-image diffusion models—renowned for high-quality synthesis—our method leverages attention manipulation \cite{hertz2022prompt, Tumanyan_2023_Plug} and LoRA fine-tuning \cite{hu2021lora}. We observe that diffusion models integrated with LoRA maintain the semantic consistency of features, enabling LoRA to learn style attributes from reference images and guide generation through carefully designed manipulation of feature and attention maps.

Based on this, we introduce mask-guided feature blending during denoising for region-specific transfer (conditioned on style references and masks) and a dynamic LoRA switching mechanism during denoising for hierarchical fusion with multiple references. These two approaches, enabling multi-LoRA combinations in spatial and temporal domains, are applicable to general diffusion methods.
Our contributions can be summed up as follows:
1) We provide new empirical insights into internal spatial features across different LoRA-integrated text-to-image diffusion models.  
2) We introduce a practical framework leveraging pre-trained and LoRA-integrated diffusion models for high-quality arbitrary image style transfer.  
3) We propose mask-guided localized style transfer and multi-style compositional style transfer for flexible model and style fusion.
4) Quantitative and qualitative results show our method outperforms state-of-the-art approaches, achieving a significantly better balance between preserving content semantics and style attributes.

\section{Related Work}
\textbf{Image Style Transfer.}
Image stylization converts images into diverse artistic styles. Originating from traditional handcrafted feature-matching between content and style images\cite{Bin2004Efficient, Zhang2013Style}, the field advanced substantially with neural style transfer\cite{Gatys_2015_Neural,liu2021adaattn,artflow2021an,tan2024style2talker}—which leverages pre-trained CNNs for feature extraction. Recent works integrate pre-trained diffusion models\cite{cui2024instastyle,10758215,wang2024instantstyle2,wang2024instantstyle1} (e.g., InST\cite{Zhang2023Inversion}, VCT\cite{cheng2023general}) to guide reference-based synthesis and style translation, while AttnD\cite{zhou2025attentiondistillationunifiedapproach}, StyleID\cite{Chung_2024_CVPR} and Z$^*$\cite{deng2023zzeroshotstyletransfer} achieve zero-shot transfer via attention manipulation. StyleShot\cite{gao2025styleshot} enables generalized transfer without test-time tuning using a style-aware encoder (decoupling training) and content-fusion encoder, and Lin et al.\cite{lin2024unsupervised} propose an unsupervised content-style learning method for cross-domain translation.

\begin{figure*}[ht]
\centering
\includegraphics[width=\linewidth]{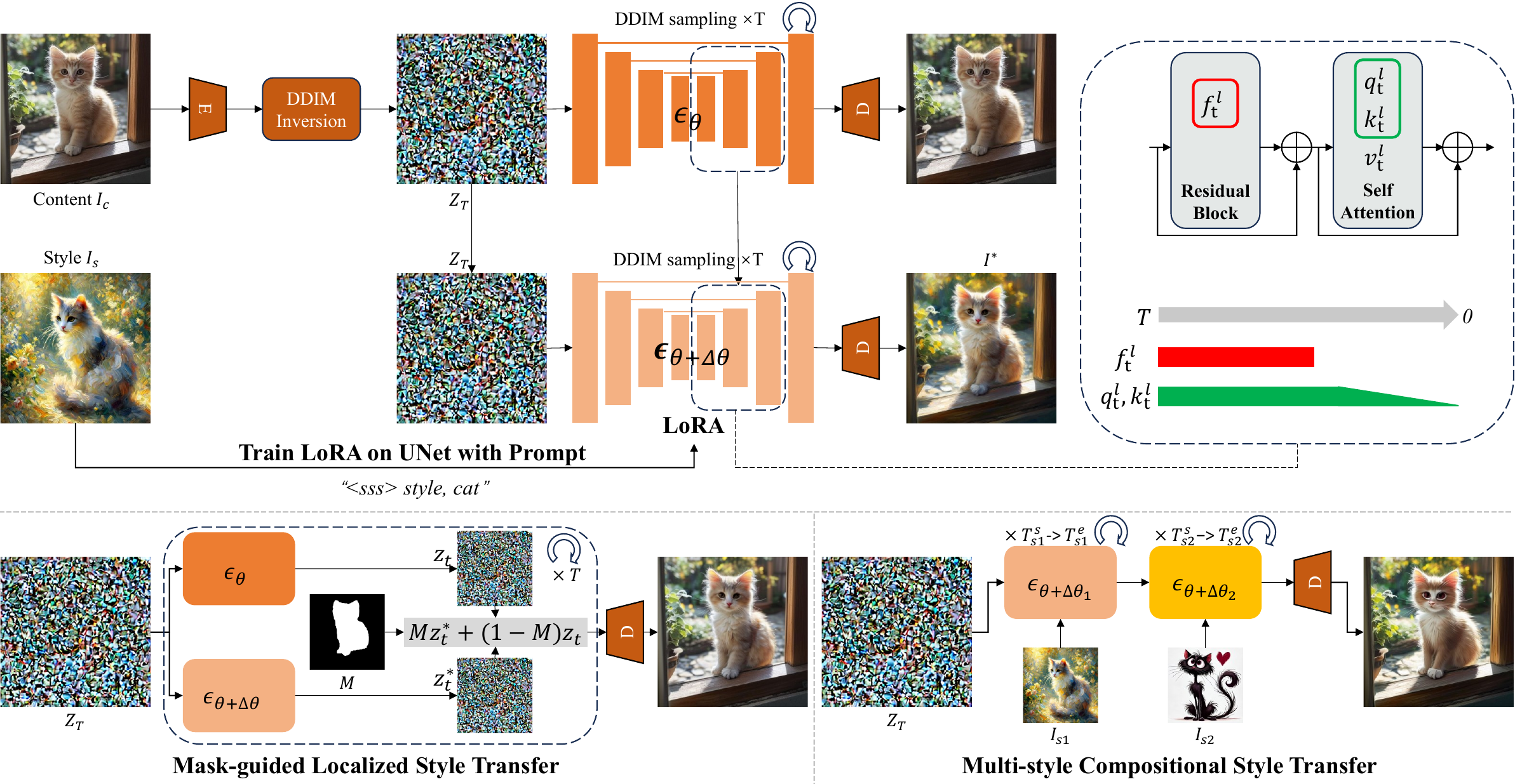}
\caption{Method pipeline. We train LoRA on UNet with a simple prompt for the style attribute transfer. To maintain the content semantics, we execute the DDIM inversion and denoising. In the DDIM denoising process, the features and attention maps are injected into the LoRA-integrated SD model. In the denoising, we override the feature in certain steps. For attention injection, we override the $q,k$ in certain steps and overlap them adaptively in the remaining steps.
For mask-guided localized style transfer, we execute the mask-guided feature combination in the denoising process.
For multi-style composition style transfer, we execute the LoRA switch in the denoising process.}
\label{fig:model}
\end{figure*}

\noindent\textbf{Semantic Localized Image Editing.}
Traditional style transfer often lacks semantic coherence, whereas text-to-image diffusion models offer flexible, controlled editing by distilling semantics from noise.  
DDIM\cite{song2022denoising} and Glide\cite{nichol2022glide} use sophisticated pixel manipulation during denoising for precise edits. Paint by Word\cite{andonian2023paint} and CLIP-inspired methods\cite{clipdiffusion,Avrahami_2022} use CLIP-guided gradients for text-image alignment in specific regions. RePaint\cite{lugmayr2022repaint} refines diffusion iterations using unmasked image info. DiffusionCLIP\cite{Kim_2022_CVPR} and SDEdit\cite{meng2022sdedit} leverage inversion/SDE priors for text-driven, realistic edits. DiffEdit\cite{couairon2022diffedit} enables targeted editing via automatic mask extraction. ILVR\cite{Choi2021ILVR} and Prompt-to-Prompt\cite{hertz2022prompt} achieve localized edits without explicit masks via attention map correlations.

\noindent\textbf{Composable Image Generation.}
Compositional image generation has drawn interest, with approaches categorized into: structured compositionality enhancement (enhancing control via scene graphs/spatial layouts\cite{Johnson2018Image,yang2022modeling,GafniMake,SinghCVPR2023High}), diffusion model adaptation (adapting generative mechanisms for compositional specs\cite{feng2023trainingfree,huang2023collaborative}), multi-concept customization (integrating multiple semantics\cite{kumari2022customdiffusion,SVDiff,gu2023mixofshow,kwon2024conceptweaverenablingmulticoncept,kong2024omgocclusionfriendlypersonalizedmulticoncept,wang2023styleadapter}), and modular model composition (composing independent models under constraints\cite{du2024reducereuserecyclecompositional,li2022composingensemblespretrainedmodels}). Notably, combining multiple stylistic attributes is underexplored, which our work addresses by investigating style target integration.

\section{Method}
This work aims to generate a new image $I^*$ that retains the semantic content of content image $I_c$ while embodying the style of style image $I_s$. Using Stable Diffusion (SD)\cite{rombach2021highresolution} as the generative backbone, we introduce techniques including DDIM inversion, LoRA training, cross-LoRA feature/attention injection, mask-guided DDIM denoising, and multi-style DDIM denoising (Fig. \ref{fig:model}).  
A key insight is that spatial features in different LoRA-integrated diffusion models exhibit strong semantic consistency. Based on this, our framework extracts features and attention maps from $I_c$'s generation process in the original SD, then injects them (along with prompts $c$) into $I_s$'s generation process in the LoRA-integrated model.  
This consistency enables multi-LoRA combined generation: we propose mask-guided localized style transfer (via feature fusion with masks) and multi-style compositional style transfer (via LoRA-switch-based sampling).

\subsection{Preliminaries}
\label{Preliminaries}
\noindent\textbf{Latent Diffusion Models.}
Diffusion models\cite{ho2020denoising,sohldickstein2015deep,song2021scorebased} represent a class of probabilistic generative models trained to reverse a diffusion process. As the latest advancement in generative models, diffusion models have garnered significant attention due to their superior capability to generate high-quality images. In the image generation, the forward diffusion process incrementally introduces noise into an initially clean image $x_0$:
\begin{equation}
\label{eq:1}
    x_t = \sqrt{\alpha_t}x_0+\sqrt{1-\alpha_t}z
\end{equation}
where $z\sim N(0,I)$ and $\{\alpha_t\}$ are the noise schedule over the time step $t$. The back process progressively removes noise from an initial Gaussian noise image. Typically, a neural network $\epsilon_\theta(x_t,t)$ is trained to predict the added noise. 

Latent Diffusion Models (LDM)\cite{rombach2021highresolution} uses a variational auto-encoder (VAE)\cite{kingma2022autoencoding} to encode the images to the latent image embeddings. A text-conditioned denoising UNet\cite{Ronneberger2015UNetCN} $\epsilon_\theta(z_t,t,c)$ is trained in the latent space. Our method is built upon Stable Diffusion (SD), a widely used pre-trained text-to-image model based on the LDM framework. A layer of the UNet in the SD involves a residual block, a self-attention block, and a cross-attention block. The attention module in UNet can be formulated as follows:
\begin{equation}
Attention(Q,K,V)=softmax(\frac{QK^T}{\sqrt{d_K}})V
    \label{eq:attention}
\end{equation}
where $Q$ denotes the query features, $K,V$ denotes the corresponding key and value feature, and $d_K$ represents the dimension of the query and key vectors. Self-attention is computed purely on the spatial features. Cross-attention is computed between spatial features and the text embeddings.

\noindent\textbf{Low-Rank Adaption (LoRA).}
Low-Rank Adaptation (LoRA)\cite{hu2021lora} is an efficient method initially employed for fine-tuning large language models by adapting a low-rank residual $\Delta\theta$, which can be decomposed into low-rank matrices, thus offering a streamlined approach to modifying the entire model parameters $\theta$. Recently, LoRA has demonstrated its effectiveness and ease of integration in fine-tuning diffusion models, particularly in personalizing text-to-image generation. Our method utilizes LoRA to transfer the attributes of the style target in image style transfer.

\begin{figure}[ht]
\centering
\includegraphics[width=\linewidth]{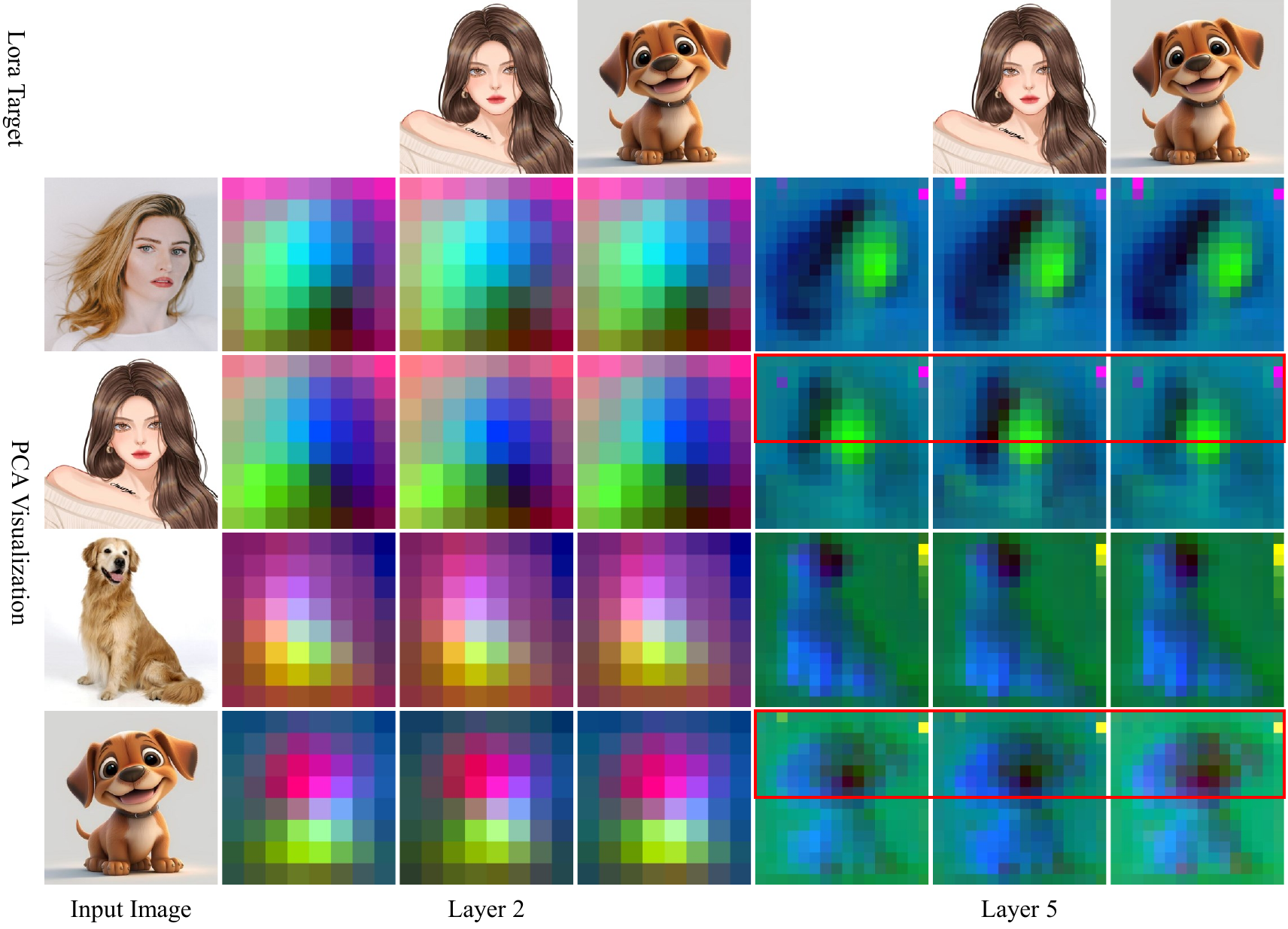}
\caption{Analysis of features from different LoRA-integrated SD: Spatial features were extracted from decoder layers at ~50\% generation (t=540). PCA on features of original SD, in-domain/out-of-domain LoRA-integrated SD (visualizing top 3 components) shows early layers focus on semantics, later on textures. LoRA-integrated models have little feature difference, with in-domain ones showing better correspondences in some layers (e.g.,layer 5).}
\label{fig:vis}
\end{figure}

\subsection{Spatial Features across LoRA Models}
We explore SD features for cross-LoRA image synthesis by fine-tuning two LoRA models in distinct stylistic domains. PCA comparisons with domain-specific images analyze features from each decoder layer at all timesteps $t$ (original SD, in-domain/out-of-domain LoRA-modified SDs). Figure \ref{fig:vis} shows initial layers capture broad semantics, later layers fine details, and LoRA-enhanced SDs have minimal feature discrepancy. Natural images exhibit indistinct model-specific feature variations, while styled images show domain-aligned patterns—specific layers (e.g., layer 5 for cartoon hair/dog faces) align more with in-domain LoRA, with out-of-domain cartoon face features reduced at layer 5.
Quantitative evaluation (20 cases, Tab. \ref{tab:lora}) confirms high feature similarity between in- and out-of-domain LoRA-modified SDs. Overall, LoRA integration subtly impacts spatial features, validating the feasibility of cross-LoRA spatial feature injection/fusion for mask-guided localized and multi-style compositional transfer.

\begin{table}[ht]
    \centering
    \setlength{\tabcolsep}{1mm}\small{\begin{tabular}{c|ccc}
    \toprule
         &Layer 2&Layer 5&Layer 7\\
         \midrule
         In-domain$\uparrow$&0.9970&0.9943&0.9922\\
         \midrule
         Out-of-domain$\uparrow$&0.9942&0.9883&0.9721\\
    \bottomrule
    \end{tabular}}
    \caption{Cosine similarity of features.}
    \label{tab:lora}
\end{table}

\subsection{Cross-LoRA Feature and Attention Injection}
Building upon the preceding observations, we propose a novel approach to style transfer. Our method learns the target's style attributes using a LoRA model, simultaneously preserving the content's semantic integrity through integrating features and attention mechanisms.

\noindent\textbf{LoRA Training.} 
Given a style image, we first train a LoRA $\Delta\theta$ on the SD UNet $\epsilon_\theta$. The learning objective for training is:
\begin{equation}
    L(\Delta\theta) = \mathbb E_{\epsilon,t}[||\epsilon-\epsilon_{\theta+\Delta\theta}(\sqrt{\alpha_t}z_0+\sqrt{1-\alpha_t}\epsilon,t,c)||]
\end{equation}
where $z_0$ denotes the encoded latent embedding of the input image, $\epsilon$ denotes the random sampled Gaussian noise, $\epsilon_{\theta+\Delta\theta}$ denotes the LoRA-integrated UNet, $c$ denotes the text embedding of the text prompt. With the LoRA-integrated SD, attributes of the style target can be injected into generating images with corresponding prompts. To avoid overfitting, the LoRA training is only carried out on the projection matrices $Q, K, V$ in the attention modules of UNet.

\noindent\textbf{Feature injection.} 
Given a content image, we invert the encoded image embedding into the initial noise using DDIM\cite{song2022denoising}. The design of shared initial noise has been proved to lead to visual similarity in the denoising\cite{Tumanyan_2023_Plug,zhang2023diffmorpher}.

At each denoising step, we extract the guidance features ${f_t^l}$ from the original denoising step: $z_{t-1}=\epsilon_\theta(z_t,t,\varnothing)$. We then inject these features into the LoRA-integrated UNet to guide the semantics of style transfer. The injection is accomplished with feature override. The operation can be expressed by:
\begin{equation}
    z_{t-1}^*=\hat{\epsilon}_{\theta+\Delta\theta}(z_t^*,t,c;\{f_t^l\})
\end{equation}
where $\hat{\epsilon}_{\theta+\Delta\theta}(\cdot;\{f_t^l\})$ denotes the denoising with injected features on the LoRA-integrated UNet. To balance the structure-preserving and the deviating, we choose layer 4 as the injection layer with a predefined injection steps threshold, following the design of Plug-and-Play\cite{Tumanyan_2023_Plug}. By doing this, a balance between 
content structure preserving and appearance deviating can be achieved.

\noindent\textbf{Attention Injection.} 
Attention injection, as a supplement to feature injection, aids in preserving details or identity consistency in image generation. In Stable Diffusion, self-attention (SA) handles spatial features, while cross-attention (CA) processes text embeddings: SA maintains spatial geometry amid style/domain variations\cite{Tumanyan_2023_Plug,chung2023style,zhang2023diffmorpher}, and CA preserves localized attributes or identities\cite{hertz2022prompt,huang2024creativesynth}.  
Layers 4-11 are chosen as injection layers to balance structure preservation and stylistic deviation.  
Fixed-threshold attention injection ensures rough structural similarity to the source, yet LoRA-induced attribute changes may cause unintended detail edits (e.g., hand deformations). To address this, we propose adaptive attention override: full attention injection is applied in early sampling steps; after threshold $T_a$, adaptive override commences.
\begin{equation}
{\{A_{t}^{l}\}^{**} } = \left\{\begin{matrix}
\{A_{t}^{l}\},&&t \ge T_a
 \\(1-\kappa_t)\{A_{t}^{l}\} + \kappa_t\{A_{t}^{l}\}^*, &&t <T_a
\end{matrix}\right.
\end{equation}
where $\{A_{t}^{l}\}$ denotes the original SD attention of the $l_{th}$ layer, $\{A_{t}^{l}\}^*$ denotes the LoRA-integrated SD attention, $T_a$ denotes the specified threshold, $\kappa_t=(T_a-t)/(T-T_a)\in[0,1]$ dynamically regulates override strength.

During each denoising step, we inject attention matrices parallel to feature injection. The modified denoising operation is formulated as follows:
\begin{equation}
    z_{t-1}^*=\hat{\epsilon}_{\theta+\Delta\theta}(z_t^*,t,c;\{f_t^l\},\{A_t^l\}^{**})
\end{equation}
Our default configuration injects self-attention into all decoder layers of the UNet architecture.

\begin{figure*}[ht]
\centering
\includegraphics[width=\linewidth]{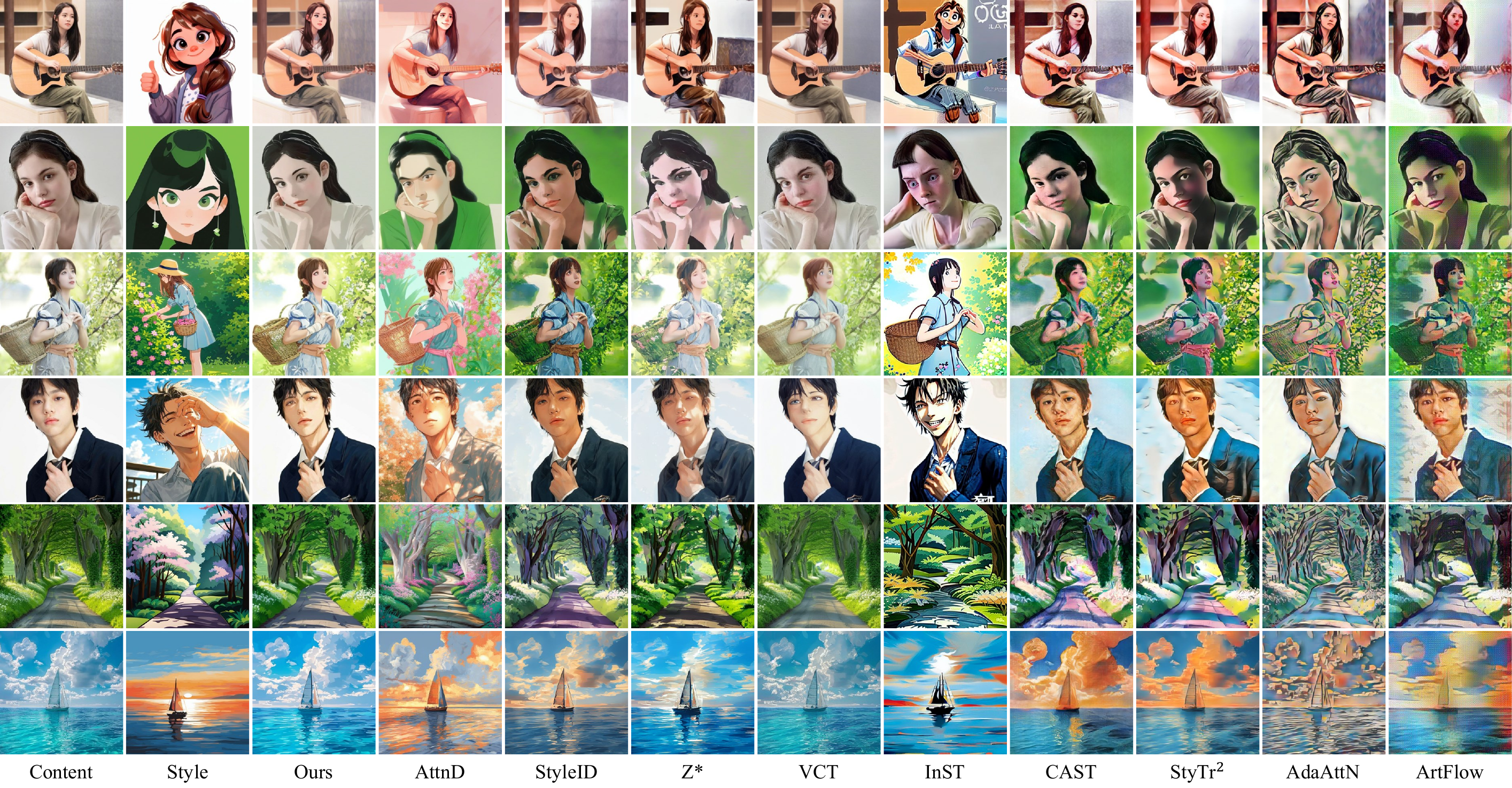}
\caption{Qualitative comparison with some state-of-the-art image style transfer methods.}
\label{fig:exp_1}
\end{figure*}

\subsection{Mask-guided Localized Image Style Transfer}
Mask-guided localized image editing focuses on modifying specific regions of an image while preserving the attributes of non-target areas. Our method involves replacing the encoded latent features within mask boundaries, guiding the denoising process through mask-based feature manipulation. Masks are resized to match the dimensions of latent features.
Mask-guided denoising is mathematically formulated as:
\begin{equation}
\hat{z}_t=Mz_t^*+(1-M)z_t.
\end{equation}
Mask-guided cross-LoRA denoising enables mask-guided style transfer. Due to the pipeline’s modular design, we can synthesize images with multiple style targets and corresponding masks:
\begin{equation}
    \begin{aligned}
\hat{z}_{(t-1)}=\sum M_i\hat{\epsilon}_{\theta+\Delta\theta_i}(\hat{z}_t,t,c;\{f_t^l\},\{A_t^l\}^{**})
\\+(1-\sum M_i){\epsilon}_{\theta}(\hat{z}_t,t,c).
    \end{aligned}
\end{equation}
where $\hat{\epsilon}_{\theta+\Delta\theta_i}$ denotes the $i_{th}$ LoRA-integrated UNet with injection, and $M_i$ denotes the corresponding region mask.

\subsection{Multi-style Composition Image Style Transfer}
We note that the UNet is capable of encoding attributes at various scales during different sampling steps. The incorporation of LoRA has a nuanced impact on spatial feature representations. Building on these findings, we introduce adaptive LoRA-switch sampling for multi-style compositional image style transfer. When we have two LoRA-integrated UNets, we can use one for generating images at a coarse-grained level and the other for a fine-grained level ($T-T_1$ steps), as shown in Fig. \ref{fig:model}. As a fundamental pipeline, multi-style LoRA-switch sampling can be extended to work with multiple (more than two) LoRA-integrated UNets, and it can be mathematically expressed as:
\begin{equation}
\hat{\epsilon}_{\theta+\Delta\theta} = \begin{matrix}
\hat{\epsilon}_{\theta+\Delta\theta_i}, &&   T_{si}^s>t\ge T_{si}^e
\end{matrix}.
\end{equation}
where $\hat{\epsilon}_{\theta+\Delta\theta_i}$ denotes the $i_{th}$ LoRA-integrated UNet with injection, and $[T_{si}^s,T_{si}^e]$ indicates the specific sampling time range for each LoRA model.

\section{Experiments}
We utilize Stable Diffusion v2.1-base as our generative backbone. Inversion and denoising are implemented using the diffusers\cite{von-platen-etal-2022-diffusers}. All experiments are performed on a single NVIDIA GeForce RTX 3090 GPU. Input and output images are processed at 512×512 resolution. Images
We train Low-Rank Adaptation (LoRA) to fine-tune the projection matrices $Q, K, V$ in the attention modules of the UNet architecture within the diffusion model. Training parameters include a LoRA rank of 16, 200 training iterations, and a learning rate of $2 \times 10^{-4}$ using the AdamW optimizer with weight decay.
Deterministic DDIM inversion and denoising are performed with 50 sampling steps derived from the full diffusion process. Our default feature and attention injection thresholds are set to 30 and 25 denoising steps out of 50 total.

\begin{table*}[ht]
    \centering
    \setlength{\tabcolsep}{1mm}\small{\begin{tabular}{c|cccccccccc}
    \toprule
         &Ours&AttnD&StyleID&$\rm Z^*$&VCT&InST&CAST&$\rm StyTr^2$&AdaAttN&ArtFlow  \\
         \midrule
         ArtFID $\downarrow$ &\textbf{22.06}& 23.62& 26.36&34.29& 36.79 & 29.41& 34.37&30.65 &29.85 & 34.79\\       
        FID $\downarrow$ &\textbf{17.85}& 19.02 & 19.76&25.58 & 28.20& 23.51& 22.23& 20.59& 20.33& 23.16 \\        
         LPIPS $\downarrow$ & \textbf{0.17}&0.18&0.27&0.29&0.26&0.20&0.48&0.42&0.40&0.44 \\
         \midrule
         CLIP $\uparrow$ &\textbf{0.36}&0.34&0.19&0.21&0.14&0.32&0.10&0.14&0.14&0.18 \\
    \bottomrule
    \end{tabular}}
    \caption{Quantitative Comparison.}
    \label{tab:my_label1}
\end{table*}

\subsection{Comparison with Style Transfer Methods}
We compare our method against leading image style transfer techniques, including five diffusion-based approaches: AttnD\cite{zhou2025attentiondistillationunifiedapproach}, StyleID\cite{Chung_2024_CVPR}, Z$^*$\cite{deng2023zzeroshotstyletransfer}, VCT\cite{cheng2023general}, and InST\cite{Zhang2023Inversion}; alongside four non-diffusion-based methods: CAST\cite{zhang2020cast}, StyTr$^2$\cite{deng2021stytr2}, AdaAttN\cite{liu2021adaattn}, and ArtFlow\cite{artflow2021an}.
Comparative results in Fig. \ref{fig:exp_1} demonstrate that diffusion-based methods excel at transferring complex stylistic attributes, including the spatial structure and visual motifs, while non-diffusion-based methods primarily alter color palettes. StyleID, Z$^*$ , and VCT preserve content properties comparable to our method, whereas AttnD and InST induce significant content alterations.
Notably, diffusion-based approaches generate high-quality, realistic outputs. While AttnD and InST produce aesthetically appealing results in certain cases, their style-content mismatches often result in artifacts like abnormal facial textures. StyleID, Z$^*$ and VCT focus more on color transformation rather than artistic style emulation.
Our model achieves superior visual quality and effectively preserves content semantics amidst stylistic transformation.

\subsubsection{Quantitative Comparison.}
Quantitative evaluation of style transfer, though challenging due to limited ground truth and large source-target style gaps, remains critical. Following prior work, we use ArtFID [(1 + LPIPS)·(1 + FID), where LPIPS quantifies content fidelity and FID assesses style fidelity] and CLIP (measuring text-image embedding similarity between outputs and style images). Evaluating 800 representative generated images (Table \ref{tab:my_label1}), our method outperforms existing approaches with the lowest ArtFID, FID, LPIPS and highest CLIP score, achieving an optimal balance between content integrity and style fidelity and setting a new benchmark.

\subsubsection{User Study.}
To compare with leading SOTA algorithms, we conducted a user study with 40 participants (18 males, 22 females, 18–40 years). They assessed 30 content-style image pairs, comparing our method with a random SOTA approach across content preservation, style capture, and overall preference. Collecting 1200 votes (Table \ref{tab:my_label}), our method received significantly more votes in all metrics, demonstrating strong balance and alignment with human perception and aesthetics, validating its efficacy and appeal.
\begin{table*}[ht]
    \centering
    \setlength{\tabcolsep}{1mm}\small\textbf{}{\begin{tabular}{c|ccccccccc}
    \toprule
         &AttnD&StyleID&$\rm Z^*$&VCT&InST&CAST&$\rm StyTr^2$&AdaAttN&ArtFlow  \\
         \midrule
         Content&37\%&21\%&23\%&45\%&26\%&37\%&32\%&42\%&37\%\\
         \midrule
         Style&44\%&29\%&37\%&28\%&41\%&17\%&16\%&15\%&14\%\\
         \midrule
         Overall&42\%&25\%&34\%&25\%&30\%&11\%&15\%&13\%&16\%\\
    \bottomrule
    \end{tabular}}
    \caption{User Study. The results show the percentage of votes where SOTA method was preferred over ours.}
    \label{tab:my_label}
\end{table*}

\begin{figure}[ht]
\centering
\includegraphics[width=\linewidth]{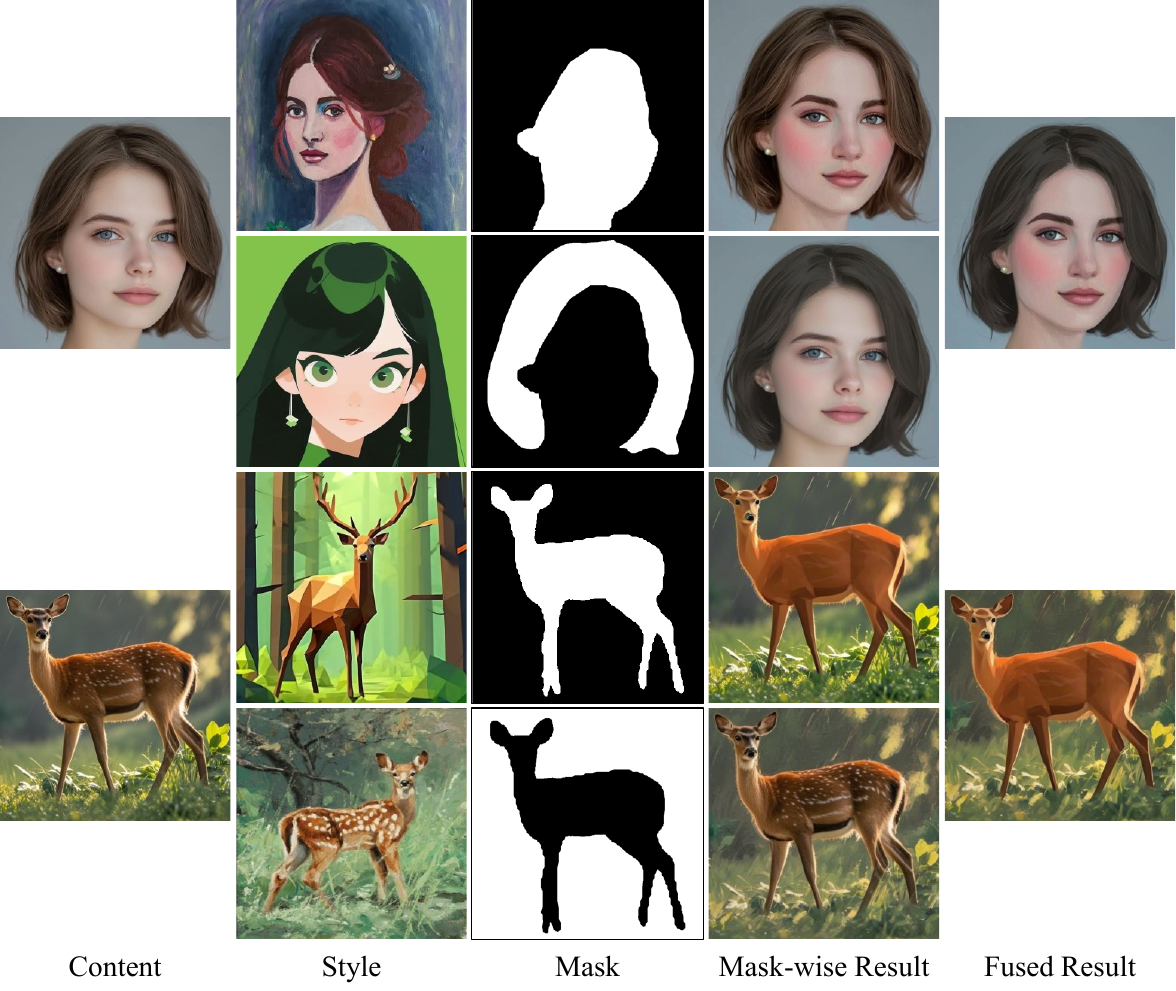}
\caption{Mask-guided localized style transfer results.}
\label{fig:exp_mask}
\end{figure}
\subsection{Mask-guided Localized Style Transfer Results}
Beyond global style transfer, our method enables mask-guided localized image style transfer, as shown in Figure \ref{fig:exp_mask} showcasing two case studies. For each content image, two style targets and corresponding masks are provided, with results including mask-guided and fused outputs. Our approach effectively synthesizes mask-guided results aligned with the mask and style target, leaving other regions unaltered. These localized results differ slightly from full style transfer outcomes.  
Comparisons reveal our method ensures harmonious generation of content and stylized regions and efficiently enables combinatorial generation of multiple styles across spatial domains.

\subsection{Multi-style Composition Style Transfer Results}
\begin{figure}[ht]
\centering
\includegraphics[width=\linewidth]{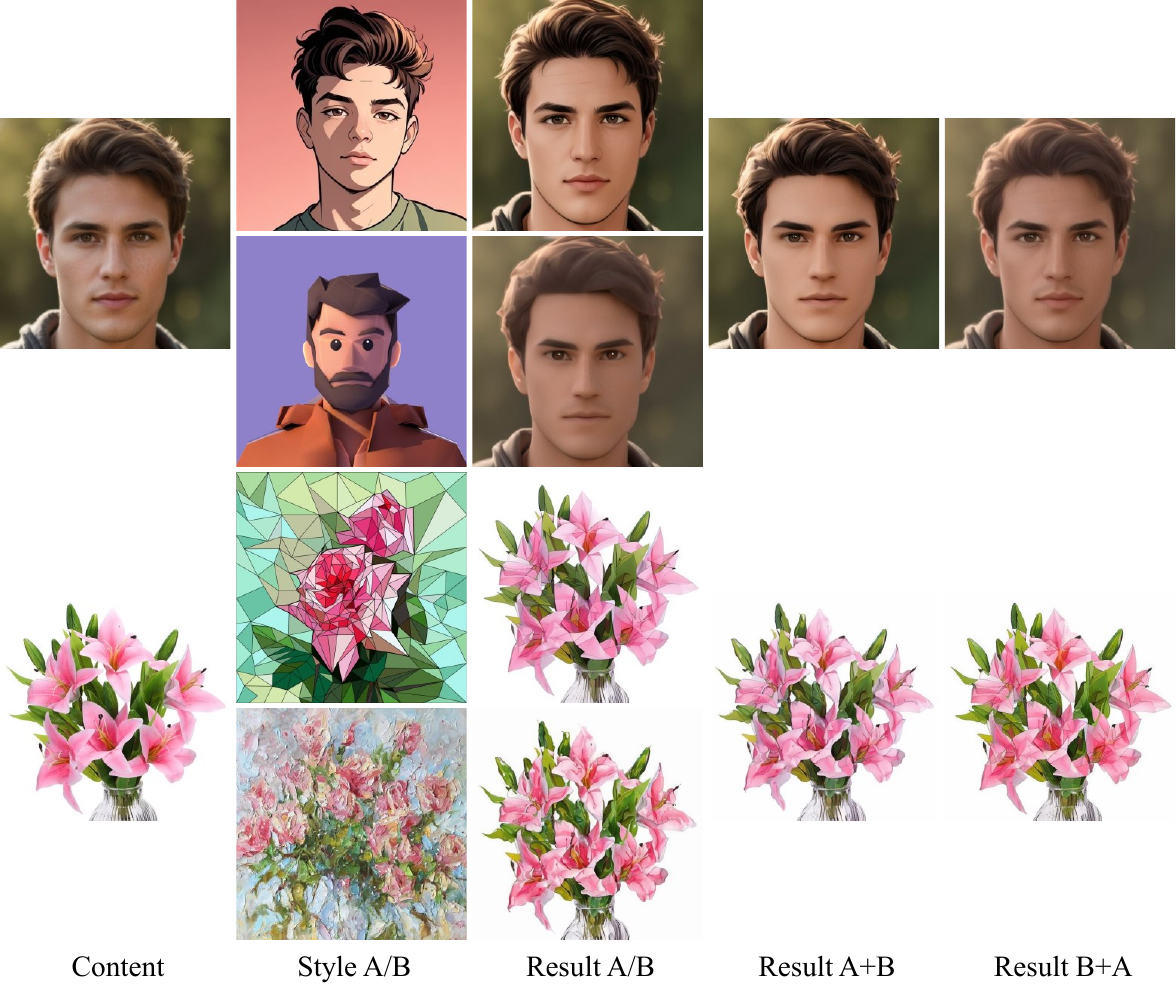}
\caption{Multi-style composition style transfer results.}
\label{fig:exp_style}
\end{figure}

We propose the multi-style composition style transfer using LoRA-switch sampling. Fig. \ref{fig:exp_style} shows two cases. For each case with two style targets $A$ and $B$, we show style transfer results of $A$, $B$, $A+B$ and $B+A$. For single-style stylization, outputs align with stylistic expectations. For style combinations, earlier sampling steps (0-30) influence global stylistic attributes while later steps (31-50) refine local details, enabling nuanced artistic expression. Our method achieves the desired stylization for any combination of two different style targets. 

\subsection{Ablation Study}
\label{Ablation}
Attention injection layers have been widely studied in prior work\cite{Tumanyan_2023_Plug,couairon2022diffedit,hertz2022prompt,zhang2023diffmorpher}. Our analysis focuses on two key aspects: cross-LoRA feature and attention injection and adaptive attention injection. These analyses validate the effectiveness of our design components. Please refer to the supplementary material for more analysis.
\begin{figure}[ht]
\centering
\includegraphics[width=\linewidth]{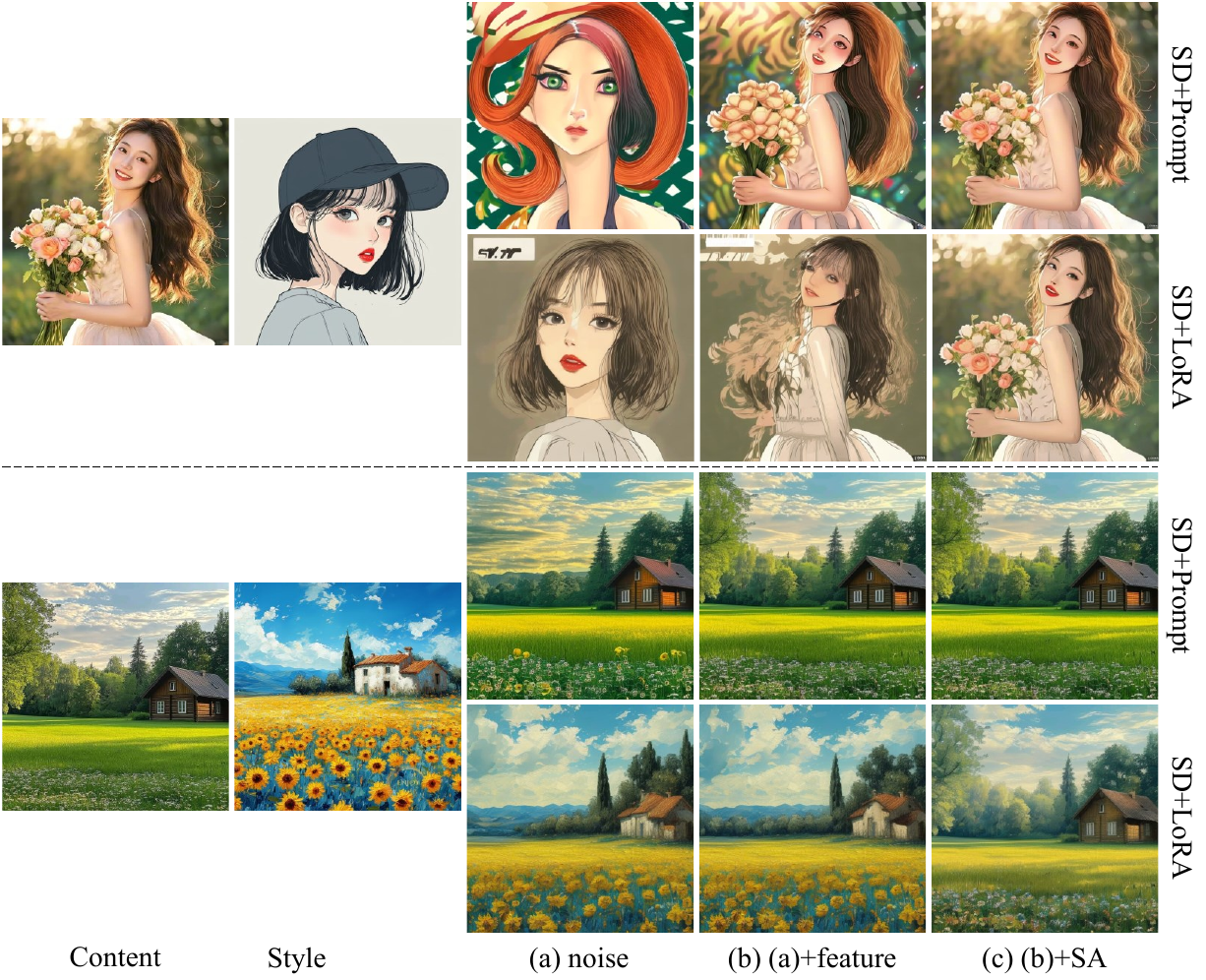}
\caption{Cross-LoRA feature and attention injection.}
\label{fig:abl1}
\end{figure}

\noindent\textbf{Cross-LoRA feature and attention injection.}
We evaluate our design via two configurations (SD+Prompt, SD+LoRA) with three synthesis combinations: (a) noise initialization, (b) (a)+feature injection, (c) (b)+self-attention (SA) injection. For SD+Prompt, generation is guided by CLIP Interrogator-optimized prompts\cite{wen2023hard}; SD+LoRA uses simplified prompts (e.g., ``$<$sss$>$ style, girl.'') for training and synthesis.  
Analyses show noise and feature injection mainly shape global spatial distribution, while SA refines local details. The two configurations yield comparable global structures but differ in local stylistic details. Notably, in portraits, noise-only inputs cause inconsistent content structures in both, yet LoRA ensures target stylistic consistency; progressive feature and attention injection enhances content consistency for more coherent stylizations. In landscapes, optimized prompts fail to fully capture target styles, whereas LoRA enables precise transfer. Our method, integrating noise initialization, feature injection, SA modulation, and LoRA adaptation, excels at preserving semantic content while transferring target stylistic attributes.

\begin{figure}[]
\centering
\includegraphics[width=\linewidth]{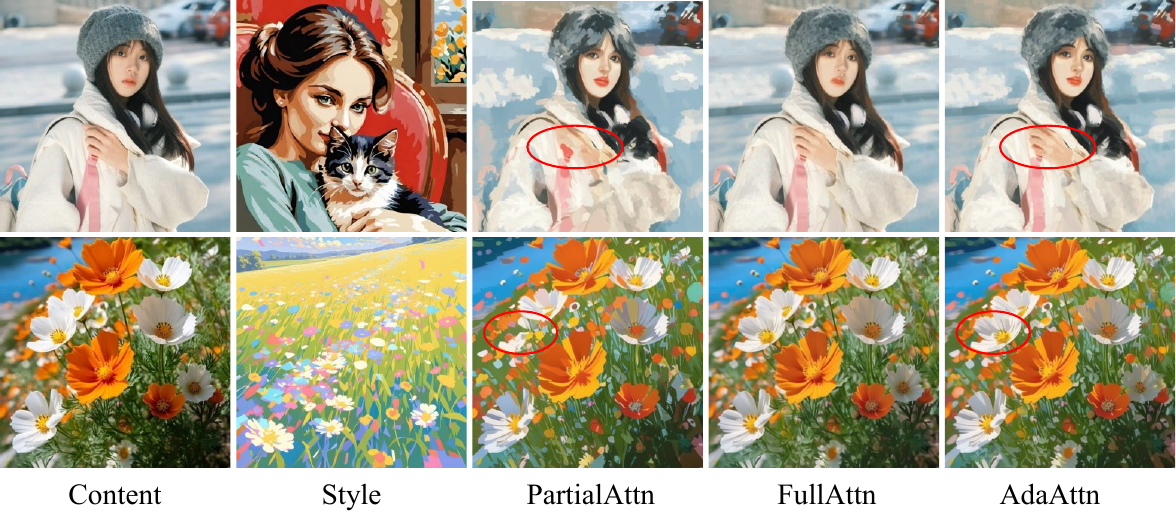}
\caption{Attn-injection with threshold 30/50 (PartialAttn), 50/50 (FullAttn) and adaptive attention injection (AdaAttn).}
\label{fig:abl_attn}
\end{figure}

\noindent\textbf{Adaptive attention injection.}
To balance content and style, we propose adaptive attention injection with an empirical threshold. Before the threshold, full attention injection enforces content guidance; beyond it, injection strength diminishes gradually. Fig. \ref{fig:abl_attn} demonstrates its effect: partial attention causes severe content loss (style leakage), full attention leads to style absence, while adaptive injection optimally balances content integrity and style fidelity.

\subsection{Discussion and Limitations}
LoRA excels at capturing nuanced target stylistic features. Confining fine-tuning to attention module projection matrices alleviates overfitting, yet challenges remain in preserving content semantic integrity: intricate style-structure entanglement hinders complete decoupling, and LoRA's absorption of stylistic properties may occasionally distort content semantics. Future work will focus on better content-style decoupling for more reliable style transfer.

Building on our approach, we propose two methods: mask-guided localized style transfer and multi-style compositional transfer. Both operate independently, leveraging Stable Diffusion's (SD) stepwise sampling independence to apply to general text-to-image models. Their use for mask-guided localized or multi-style compositional text-to-image generation yields desired results, with further analysis in the Supplementary Material.

\section{Conclusion}
In this paper, we introduce a novel style transfer framework that captures stylistic attributes from a single reference image via LoRA fine-tuning and guides image synthesis through feature and attention injection. The framework is rooted in a core insight: the inherent cross-LoRA semantic consistency within text-to-image diffusion models. Our approach outperforms state-of-the-art methods by achieving a superior balance between preserving content semantics and transferring stylistic attributes. Building on this insight, we further introduce two innovative techniques: mask-guided localized style transfer and multi-style compositional style transfer. Comprehensive experimental evaluations demonstrate that our approach surpasses existing SOTA techniques.

\section{Acknowledgement}
This work is supported by National Natural Science Foundation of China (NSFC, No. 62472285 and No. 62102255), Sponsored by CCF-NetEase ThunderFire Innovation Research Funding (NO. CCF-Netease 202508 and NO. CCF-Netease 202509).

\bibliography{aaai2026}

@article{hertz2022prompt,
  title={Prompt-to-prompt image editing with cross attention control},
  author={Hertz, Amir and Mokady, Ron and Tenenbaum, Jay and Aberman, Kfir and Pritch, Yael and Cohen-Or, Daniel},
  booktitle={arXiv preprint arXiv:2208.01626},
  year={2022}
}

@misc{huang2024creativesynth,
      title={CreativeSynth: Creative Blending and Synthesis of Visual Arts based on Multimodal Diffusion}, 
      author={Nisha Huang and Weiming Dong and Yuxin Zhang and Fan Tang and Ronghui Li and Chongyang Ma and Xiu Li and Changsheng Xu},
      year={2024},
      eprint={2401.14066},
      archivePrefix={arXiv},
      primaryClass={cs.CV}
}

@article{chung2023style,
  title={Style Injection in Diffusion: A Training-free Approach for Adapting Large-scale Diffusion Models for Style Transfer},
  author={Chung, Jiwoo and Hyun, Sangeek and Heo, Jae-Pil},
  journal={arXiv preprint arXiv:2312.09008},
  year={2023}
}

@INPROCEEDINGS{Zhang2023Inversion,
  author={Zhang, Yuxin and Huang, Nisha and Tang, Fan and Huang, Haibin and Ma, Chongyang and Dong, Weiming and Xu, Changsheng},
  booktitle={CVPR}, 
  title={Inversion-based Style Transfer with Diffusion Models}, 
  year={2023},
  volume={},
  number={},
  pages={10146-10156},
  doi={10.1109/CVPR52729.2023.00978}}

@misc{jeong2024visual,
      title={Visual Style Prompting with Swapping Self-Attention}, 
      author={Jaeseok Jeong and Junho Kim and Yunjey Choi and Gayoung Lee and Youngjung Uh},
      year={2024},
      eprint={2402.12974},
      archivePrefix={arXiv},
      primaryClass={cs.CV}}

@article{cheng2023general,
  title={General Image-to-Image Translation with One-Shot Image Guidance},
  author={Cheng, B. and Liu, Z. and Peng, Y. and Lin, Y.},
  journal={arXiv preprint arXiv:2307.14352},
  year={2023}
}

@inproceedings{zhang2020cast,
author = {Zhang, Yuxin and Tang, Fan and Dong, Weiming and Huang, Haibin and Ma, Chongyang and Lee, Tong-Yee and Xu, Changsheng},
title = {Domain Enhanced Arbitrary Image Style Transfer via Contrastive Learning},
booktitle = {ACM SIGGRAPH},
year = {2022}}

@inproceedings{liu2021adaattn,
  title={AdaAttN: Revisit Attention Mechanism in Arbitrary Neural Style Transfer},
  author={Liu, Songhua and Lin, Tianwei and He, Dongliang and Li, Fu and Wang, Meiling and Li, Xin and Sun, Zhengxing and Li, Qian and Ding, Errui},
  booktitle={Proceedings of the IEEE International Conference on Computer Vision},
  year={2021}
}

@inproceedings{artflow2021an,
 title={ArtFlow: Unbiased image style transfer via reversible neural flows},
 author={An, Jie and Huang, Siyu and Song, Yibing and Dou, Dejing and Liu, Wei and Luo, Jiebo},
 booktitle={Proceedings of the IEEE/CVF Conference on Computer Vision and Pattern Recognition},
 year={2021}
}

@inproceedings{deng2021stytr2,
      title={StyTr $^2$: Image Style Transfer with Transformers}, 
      author={Yingying Deng and Fan Tang and Weiming Dong and Chongyang Ma and Xingjia Pan and Lei Wang and Changsheng Xu},
      booktitle={CVPR},
      year={2022},
}

@misc{von-platen-etal-2022-diffusers,
  author = {Patrick von Platen and Suraj Patil and Anton Lozhkov and Pedro Cuenca and Nathan Lambert and Kashif Rasul and Mishig Davaadorj and Dhruv Nair and Sayak Paul and William Berman and Yiyi Xu and Steven Liu and Thomas Wolf},
  title = {Diffusers: State-of-the-art diffusion models},
  year = {2022},
  publisher = {GitHub},
  journal = {GitHub repository},
  howpublished = {\url{https://github.com/huggingface/diffusers}}
}

@misc{rombach2021highresolution,
      title={High-Resolution Image Synthesis with Latent Diffusion Models}, 
      author={Robin Rombach and Andreas Blattmann and Dominik Lorenz and Patrick Esser and Björn Ommer},
      year={2021},
      eprint={2112.10752},
      archivePrefix={arXiv},
      primaryClass={cs.CV}
}

@misc{hu2021lora,
      title={LoRA: Low-Rank Adaptation of Large Language Models}, 
      author={Edward J. Hu and Yelong Shen and Phillip Wallis and Zeyuan Allen-Zhu and Yuanzhi Li and Shean Wang and Lu Wang and Weizhu Chen},
      year={2021},
      eprint={2106.09685},
      archivePrefix={arXiv},
      primaryClass={cs.CL}
}

@misc{song2022denoising,
      title={Denoising Diffusion Implicit Models}, 
      author={Jiaming Song and Chenlin Meng and Stefano Ermon},
      year={2022},
      eprint={2010.02502},
      archivePrefix={arXiv},
      primaryClass={cs.LG}
}

@misc{gal2022textual,
      doi = {10.48550/ARXIV.2208.01618},
      url = {https://arxiv.org/abs/2208.01618},
      author = {Gal, Rinon and Alaluf, Yuval and Atzmon, Yuval and Patashnik, Or and Bermano, Amit H. and Chechik, Gal and Cohen-Or, Daniel},
      title = {An Image is Worth One Word: Personalizing Text-to-Image Generation using Textual Inversion},
      publisher = {arXiv},
      year = {2022},
      primaryClass={cs.CV}
}

@ARTICLE{Bin2004Efficient,
  author={Bin Wang and Wenping Wang and Huaiping Yang and Jiaguang Sun},
  journal={IEEE Transactions on Visualization and Computer Graphics}, 
  title={Efficient example-based painting and synthesis of 2D directional texture}, 
  year={2004},
  volume={10},
  number={3},
  pages={266-277},
  doi={10.1109/TVCG.2004.1272726}}

@ARTICLE{Zhang2013Style,
  author={Zhang, Wei and Cao, Chen and Chen, Shifeng and Liu, Jianzhuang and Tang, Xiaoou},
  journal={IEEE Transactions on Multimedia}, 
  title={Style Transfer Via Image Component Analysis}, 
  year={2013},
  volume={15},
  number={7},
  pages={1594-1601},
  doi={10.1109/TMM.2013.2265675}}

@INPROCEEDINGS{Choi2021ILVR,
  author={Choi, Jooyoung and Kim, Sungwon and Jeong, Yonghyun and Gwon, Youngjune and Yoon, Sungroh},
  booktitle={2021 IEEE/CVF International Conference on Computer Vision (ICCV)}, 
  title={ILVR: Conditioning Method for Denoising Diffusion Probabilistic Models}, 
  year={2021},
  volume={},
  number={},
  pages={14347-14356},
  doi={10.1109/ICCV48922.2021.01410}}

@misc{ramesh2022hierarchical,
      title={Hierarchical Text-Conditional Image Generation with CLIP Latents}, 
      author={Aditya Ramesh and Prafulla Dhariwal and Alex Nichol and Casey Chu and Mark Chen},
      year={2022},
      eprint={2204.06125},
      archivePrefix={arXiv},
      primaryClass={cs.CV}
}

@misc{nichol2022glide,
      title={GLIDE: Towards Photorealistic Image Generation and Editing with Text-Guided Diffusion Models}, 
      author={Alex Nichol and Prafulla Dhariwal and Aditya Ramesh and Pranav Shyam and Pamela Mishkin and Bob McGrew and Ilya Sutskever and Mark Chen},
      year={2022},
      eprint={2112.10741},
      archivePrefix={arXiv},
      primaryClass={cs.CV}
}

@misc{andonian2023paint,
      title={Paint by Word}, 
      author={Alex Andonian and Sabrina Osmany and Audrey Cui and YeonHwan Park and Ali Jahanian and Antonio Torralba and David Bau},
      year={2023},
      eprint={2103.10951},
      archivePrefix={arXiv},
      primaryClass={cs.CV}
}

@inproceedings{Avrahami_2022,
   title={Blended Diffusion for Text-driven Editing of Natural Images},
   url={http://dx.doi.org/10.1109/CVPR52688.2022.01767},
   DOI={10.1109/cvpr52688.2022.01767},
   booktitle={CVPR},
   publisher={IEEE},
   author={Avrahami, Omri and Lischinski, Dani and Fried, Ohad},
   year={2022},
   month={jun}, }

@misc{clipdiffusion,
  author = {Katherine Crowson},
  title = {CLIP-Guided-Diffusion},
  howpublished ={\url{https://colab.research.google.com/drive/1V66mUeJbXrTuQITvJunvnWVn96FEbSI3.}},
  year = 2023,
}

@misc{lugmayr2022repaint,
      title={RePaint: Inpainting using Denoising Diffusion Probabilistic Models}, 
      author={Andreas Lugmayr and Martin Danelljan and Andres Romero and Fisher Yu and Radu Timofte and Luc Van Gool},
      year={2022},
      eprint={2201.09865},
      archivePrefix={arXiv},
      primaryClass={cs.CV}
}

@InProceedings{Kim_2022_CVPR,
    author    = {Kim, Gwanghyun and Kwon, Taesung and Ye, Jong Chul},
    title     = {DiffusionCLIP: Text-Guided Diffusion Models for Robust Image Manipulation},
    booktitle = {CVPR},
    month     = {June},
    year      = {2022},
    pages     = {2426-2435}
}

@misc{couairon2022diffedit,
      title={DiffEdit: Diffusion-based semantic image editing with mask guidance}, 
      author={Guillaume Couairon and Jakob Verbeek and Holger Schwenk and Matthieu Cord},
      year={2022},
      eprint={2210.11427},
      archivePrefix={arXiv},
      primaryClass={cs.CV}
}

@article{Gatys_2015_Neural,
  title={A Neural Algorithm of Artistic Style},
  author={Gatys, Leon A and Ecker, Alexander S and Bethge, Matthias},
  journal={arXiv preprint arXiv:1508.06576},
  year={2015}
}

@InProceedings{Chung_2024_CVPR,
    author    = {Chung, Jiwoo and Hyun, Sangeek and Heo, Jae-Pil},
    title     = {Style Injection in Diffusion: A Training-free Approach for Adapting Large-scale Diffusion Models for Style Transfer},
    booktitle = {CVPR},
    month     = {June},
    year      = {2024},
    pages     = {8795-8805}
}

@misc{deng2023zzeroshotstyletransfer,
      title={$Z^*$: Zero-shot Style Transfer via Attention Rearrangement}, 
      author={Yingying Deng and Xiangyu He and Fan Tang and Weiming Dong},
      year={2023},
      eprint={2311.16491},
      archivePrefix={arXiv},
      primaryClass={cs.CV},
      url={https://arxiv.org/abs/2311.16491}, 
}

@misc{zhou2025attentiondistillationunifiedapproach,
      title={Attention Distillation: A Unified Approach to Visual Characteristics Transfer}, 
      author={Yang Zhou and Xu Gao and Zichong Chen and Hui Huang},
      year={2025},
      eprint={2502.20235},
      archivePrefix={arXiv},
      primaryClass={cs.CV},
      url={https://arxiv.org/abs/2502.20235}, 
}

@INPROCEEDINGS{Johnson2018Image,
  author={Johnson, Justin and Gupta, Agrim and Fei-Fei, Li},
  booktitle={2018 IEEE/CVF Conference on Computer Vision and Pattern Recognition}, 
  title={Image Generation from Scene Graphs}, 
  year={2018},
  volume={},
  number={},
  pages={1219-1228},
  doi={10.1109/CVPR.2018.00133}}

@inproceedings{yang2022modeling,
  title={Modeling image composition for complex scene generation},
  author={Yang, Zuopeng and Liu, Daqing and Wang, Chaoyue and Yang, Jie and Tao, Dacheng},
  booktitle={Proceedings of the IEEE/CVF Conference on Computer Vision and Pattern Recognition},
  pages={7764--7773},
  year={2022}
}

@misc{GafniMake,
  doi = {10.48550/ARXIV.2203.13131},
  url = {https://arxiv.org/abs/2203.13131},
  author = {Gafni, Oran and Polyak, Adam and Ashual, Oron and Sheynin, Shelly and Parikh, Devi and Taigman, Yaniv},
  title = {Make-A-Scene: Scene-Based Text-to-Image Generation with Human Priors},
  publisher = {arXiv},
  year = {2022},
  copyright = {arXiv.org perpetual, non-exclusive license}
}

@inproceedings{feng2023trainingfree,
title={Training-Free Structured Diffusion Guidance for Compositional Text-to-Image Synthesis},
author={Weixi Feng and Xuehai He and Tsu-Jui Fu and Varun Jampani and Arjun Reddy Akula and Pradyumna Narayana and Sugato Basu and Xin Eric Wang and William Yang Wang},
booktitle={The Eleventh International Conference on Learning Representations },
year={2023},
url={https://openreview.net/forum?id=PUIqjT4rzq7}
}

@InProceedings{huang2023collaborative,
     author = {Huang, Ziqi and Chan, Kelvin C.K. and Jiang, Yuming and Liu, Ziwei},
     title = {Collaborative Diffusion for Multi-Modal Face Generation and Editing},
     booktitle = {Proceedings of the IEEE/CVF Conference on Computer Vision and Pattern Recognition},
     year = {2023},
 }

@article{kumari2022customdiffusion,
  title={Multi-Concept Customization of Text-to-Image Diffusion},
  author={Kumari, Nupur and Zhang, Bingliang and Zhang, Richard and Shechtman, Eli and Zhu, Jun-Yan},
  booktitle = {Proceedings of the IEEE/CVF Conference on Computer Vision and Pattern Recognition (CVPR)},
  year      = {2023}
}

@misc{SVDiff,
      title         = {SVDiff: Compact Parameter Space for Diffusion Fine-Tuning}, 
      author        = {Ligong Han and Yinxiao Li and Han Zhang and Peyman Milanfar and Dimitris Metaxas and Feng Yang},
      year          = {2023},
      eprint        = {2303.11305},
      archivePrefix = {arXiv},
      primaryClass  = {cs.CV},
      url           = {https://arxiv.org/abs/2303.11305}
}

@article{gu2023mixofshow,
    title={Mix-of-Show: Decentralized Low-Rank Adaptation for Multi-Concept Customization of Diffusion Models},
    author={Gu, Yuchao and Wang, Xintao and Wu, Jay Zhangjie and Shi, Yujun and Chen Yunpeng and Fan, Zihan and Xiao, Wuyou and Zhao, Rui and Chang, Shuning and Wu, Weijia and Ge, Yixiao and Shan Ying and Shou, Mike Zheng},
    journal={arXiv preprint arXiv:2305.18292},
    year={2023}
}

@misc{kwon2024conceptweaverenablingmulticoncept,
      title={Concept Weaver: Enabling Multi-Concept Fusion in Text-to-Image Models}, 
      author={Gihyun Kwon and Simon Jenni and Dingzeyu Li and Joon-Young Lee and Jong Chul Ye and Fabian Caba Heilbron},
      year={2024},
      eprint={2404.03913},
      archivePrefix={arXiv},
      primaryClass={cs.CV},
      url={https://arxiv.org/abs/2404.03913}, 
}

@misc{kong2024omgocclusionfriendlypersonalizedmulticoncept,
      title={OMG: Occlusion-friendly Personalized Multi-concept Generation in Diffusion Models}, 
      author={Zhe Kong and Yong Zhang and Tianyu Yang and Tao Wang and Kaihao Zhang and Bizhu Wu and Guanying Chen and Wei Liu and Wenhan Luo},
      year={2024},
      eprint={2403.10983},
      archivePrefix={arXiv},
      primaryClass={cs.CV},
      url={https://arxiv.org/abs/2403.10983}, 
}

@misc{du2024reducereuserecyclecompositional,
      title={Reduce, Reuse, Recycle: Compositional Generation with Energy-Based Diffusion Models and MCMC}, 
      author={Yilun Du and Conor Durkan and Robin Strudel and Joshua B. Tenenbaum and Sander Dieleman and Rob Fergus and Jascha Sohl-Dickstein and Arnaud Doucet and Will Grathwohl},
      year={2024},
      eprint={2302.11552},
      archivePrefix={arXiv},
      primaryClass={cs.LG},
      url={https://arxiv.org/abs/2302.11552}, 
}

@misc{li2022composingensemblespretrainedmodels,
      title={Composing Ensembles of Pre-trained Models via Iterative Consensus}, 
      author={Shuang Li and Yilun Du and Joshua B. Tenenbaum and Antonio Torralba and Igor Mordatch},
      year={2022},
      eprint={2210.11522},
      archivePrefix={arXiv},
      primaryClass={cs.CV},
      url={https://arxiv.org/abs/2210.11522}, 
}

@INPROCEEDINGS{SinghCVPR2023High,
  author={Singh, Jaskirat and Gould, Stephen and Zheng, Liang},
  booktitle={2023 IEEE/CVF Conference on Computer Vision and Pattern Recognition (CVPR)}, 
  title={High-Fidelity Guided Image Synthesis with Latent Diffusion Models}, 
  year={2023},
  volume={},
  number={},
  pages={5997-6006},
  doi={10.1109/CVPR52729.2023.00581}}

@article{wang2023styleadapter,
  title={StyleAdapter: A Unified Stylized Image Generation Model},
  author={Wang, Zhouxia and Wang, Xintao and Xie, Liangbin and Qi, Zhongang and Shan, Ying and Wang, Wenping and Luo, Ping},
  journal={arXiv preprint arXiv:2309.01770},
  year={2023}
}

@article{lv2024color,
  title={Color transfer for images: A survey},
  author={Lv, Chenlei and Zhang, Dan and Geng, Shengling and Wu, Zhongke and Huang, Hui},
  journal={ACM Transactions on Multimedia Computing, Communications and Applications},
  volume={20},
  number={8},
  pages={1--29},
  year={2024},
  publisher={ACM New York, NY}
}

@article{zhou2024comprehensive,
  title={A Comprehensive Evaluation of Arbitrary Image Style Transfer Methods},
  author={Zhou, Zijun and Tang, Fan and Zhang, Yuxin and Deussen, Oliver and Cao, Juan and Dong, Weiming and Li, Xiangtao and Lee, Tong-Yee},
  journal={IEEE Transactions on Visualization and Computer Graphics},
  year={2024},
  publisher={IEEE}
}

@inproceedings{cui2024instastyle,
  title={InstaStyle: Inversion Noise of a Stylized Image is Secretly a Style Adviser},
  author={Cui, Xing and Li, Zekun and Li, Pei Pei and Huang, Huaibo and Liu, Xuannan and He, Zhaofeng},
  booktitle={ECCV},
  year={2024}}

@ARTICLE{10758215,
  author={Yang, Rui and Wu, Xiaojun and He, Shengfeng},
  journal={IEEE Transactions on Visualization and Computer Graphics}, 
  title={MixSA: Training-free Reference-based Sketch Extraction via Mixture-of-Self-Attention}, 
  year={2024},
  volume={},
  number={},
  pages={1-16},
  doi={10.1109/TVCG.2024.3502395}}

@article{wang2024instantstyle1,
  title={InstantStyle-Plus: Style Transfer with Content-Preserving in Text-to-Image Generation},
  author={Wang, Haofan and Xing, Peng and Huang, Renyuan and Ai, Hao and Wang, Qixun and Bai, Xu},
  journal={arXiv preprint arXiv:2407.00788},
  year={2024}
}

@article{wang2024instantstyle2,
  title={InstantStyle: Free Lunch towards Style-Preserving in Text-to-Image Generation},
  author={Wang, Haofan and Wang, Qixun and Bai, Xu and Qin, Zekui and Chen, Anthony},
  journal={arXiv preprint arXiv:2404.02733},
  year={2024}
}

@article{wen2023hard,
  title={Hard prompts made easy: Gradient-based discrete optimization for prompt tuning and discovery},
  author={Wen, Yuxin and Jain, Neel and Kirchenbauer, John and Goldblum, Micah and Geiping, Jonas and Goldstein, Tom},
  journal={Advances in Neural Information Processing Systems},
  volume={36},
  pages={51008--51025},
  year={2023}
}

@article{zhang2023diffmorpher,
    title={DiffMorpher: Unleashing the Capability of Diffusion Models for Image Morphing},
    author={Zhang, Kaiwen and Zhou, Yifan and Xu, Xudong and Pan, Xingang and Dai, Bo},
    journal={arXiv preprint arXiv:2312.07409},
    year={2023}
}

@misc{ho2020denoising,
      title={Denoising Diffusion Probabilistic Models}, 
      author={Jonathan Ho and Ajay Jain and Pieter Abbeel},
      year={2020},
      eprint={2006.11239},
      archivePrefix={arXiv},
      primaryClass={cs.LG}
}

@misc{sohldickstein2015deep,
      title={Deep Unsupervised Learning using Nonequilibrium Thermodynamics}, 
      author={Jascha Sohl-Dickstein and Eric A. Weiss and Niru Maheswaranathan and Surya Ganguli},
      year={2015},
      eprint={1503.03585},
      archivePrefix={arXiv},
      primaryClass={cs.LG}
}

@inproceedings{
  song2021scorebased,
  title={Score-Based Generative Modeling through Stochastic Differential Equations},
  author={Yang Song and Jascha Sohl-Dickstein and Diederik P Kingma and Abhishek Kumar and Stefano Ermon and Ben Poole},
  booktitle={ICLR},
  year={2021},
}

@misc{kingma2022autoencoding,
      title={Auto-Encoding Variational Bayes}, 
      author={Diederik P Kingma and Max Welling},
      year={2022},
      eprint={1312.6114},
      archivePrefix={arXiv},
      primaryClass={stat.ML}
}

@article{Ronneberger2015UNetCN,
  title={U-Net: Convolutional Networks for Biomedical Image Segmentation},
  author={Olaf Ronneberger and Philipp Fischer and Thomas Brox},
  journal={ArXiv},
  year={2015},
  volume={abs/1505.04597},
}

@InProceedings{Tumanyan_2023_Plug,
    author    = {Tumanyan, Narek and Geyer, Michal and Bagon, Shai and Dekel, Tali},
    title     = {Plug-and-Play Diffusion Features for Text-Driven Image-to-Image Translation},
    booktitle = {CVPR},
    month     = {June},
    year      = {2023},
    pages     = {1921-1930}
}

@inproceedings{
      meng2022sdedit,
      title={{SDE}dit: Guided Image Synthesis and Editing with Stochastic Differential Equations},
      author={Chenlin Meng and Yutong He and Yang Song and Jiaming Song and Jiajun Wu and Jun-Yan Zhu and Stefano Ermon},
      booktitle={ICLR},
      year={2022},
}

@inproceedings{tan2024style2talker,
  title={Style2Talker: High-Resolution Talking Head Generation with Emotion Style and Art Style},
  author={Tan, Shuai and Ji, Bin and Pan, Ye},
  booktitle={Proceedings of the AAAI Conference on Artificial Intelligence},
  volume={38},
  number={5},
  pages={5079--5087},
  year={2024}
}

@article{gao2025styleshot,
  title={Styleshot: A snapshot on any style},
  author={Gao, Junyao and Sun, Yanan and Liu, Yanchen and Tang, Yinhao and Zeng, Yanhong and Qi, Ding and Chen, Kai and Zhao, Cairong},
  journal={IEEE Transactions on Pattern Analysis and Machine Intelligence},
  year={2025},
  publisher={IEEE}
}

@article{lin2024unsupervised,
  title={Unsupervised content and style learning for multimodal cross-domain image translation},
  author={Lin, Zhijie and Chen, Jingjing and Ma, Xiaolong and Li, Chao and Zhang, Huiming and Zhao, Lei},
  journal={Scientific Reports},
  volume={14},
  number={1},
  pages={29469},
  year={2024},
  publisher={Nature Publishing Group UK London}
}

\end{document}